\documentclass[sigconf]{acmart}

\copyrightyear{2025}
\acmYear{2025}
\setcopyright{cc}
\setcctype{by}
\acmConference[UbiComp Companion '25]{Companion of the 2025 ACM International Joint Conference on Pervasive and Ubiquitous Computing}{October 12--16, 2025}{Espoo, Finland}
\acmBooktitle{Companion of the 2025 ACM International Joint Conference on Pervasive and Ubiquitous Computing (UbiComp Companion '25), October 12--16, 2025, Espoo, Finland}\acmDOI{10.1145/3714394.3754412}
\acmISBN{979-8-4007-1477-1/2025/10}

\usepackage{algorithm}
\usepackage[noend]{algpseudocode} 
\algrenewcommand\algorithmiccomment[1]{\hfill$\triangleright$~#1}
\algrenewcommand\algorithmicrequire{\textbf{Input:}}
\algrenewcommand\algorithmicensure{\textbf{Output:}}

\begin{document}

\title[Motion-Robust Multimodal Fusion of PPG and Accelerometer Signals \\ for Three-Class Heart Rhythm Classification]{Motion-Robust Multimodal Fusion of PPG and Accelerometer Signals for Three-Class Heart Rhythm Classification}

\author{Yangyang Zhao}
\email{yazhao@utu.fi}
\affiliation{
  \institution{Department of Computing, Faculty of Technology, University of Turku}
  \city{Turku}
  \postcode{20520}
  \country{Finland}
}

\author{Matti Kaisti}
\email{mkaist@utu.fi}
\affiliation{
  \institution{Department of Computing, Faculty of Technology, University of Turku}
  \city{Turku}
  \postcode{20520}
  \country{Finland}
}

\author{Olli Lahdenoja}
\email{olanla@utu.fi}
\affiliation{
  \institution{Department of Computing, Faculty of Technology, University of Turku}
  \city{Turku}
  \postcode{20520}
  \country{Finland}
}

\author{Tero Koivisto}
\email{tejuko@utu.fi}
\affiliation{
  \institution{Department of Computing, Faculty of Technology, University of Turku}
  \city{Turku}
  \postcode{20520}
  \country{Finland}
}

\renewcommand{\shortauthors}{Yangyang Zhao, Matti Kaisti, Olli Lahdenoja, \& Tero Koivisto}

\begin{abstract}

Atrial fibrillation (AF) is a leading cause of stroke and mortality, particularly in elderly patients. Wrist-worn photoplethysmography (PPG) enables non-invasive, continuous rhythm monitoring, yet suffers from significant vulnerability to motion artifacts and physiological noise. Many existing approaches rely solely on single-channel PPG and are limited to binary AF detection, often failing to capture the broader range of arrhythmias encountered in clinical settings.
We introduce RhythmiNet, a residual neural network enhanced with temporal and channel attention modules that jointly leverage PPG and accelerometer (ACC) signals. The model performs three-class rhythm classification: AF, sinus rhythm (SR), and Other. To assess robustness across varying movement conditions, test data are stratified by accelerometer-based motion intensity percentiles without excluding any segments.
RhythmiNet achieved a 4.3\% improvement in macro-AUC over the PPG-only baseline. In addition, performance surpassed a logistic regression model based on handcrafted HRV features by 12\%, highlighting the benefit of multimodal fusion and attention-based learning in noisy, real-world clinical data.
\end{abstract}



\maketitle

\section{INTRODUCTION}
Atrial fibrillation (AF) is the most prevalent sustained arrhythmia and a major contributor to stroke, heart failure, and mortality, particularly in elderly populations \cite{wesselius2021digital}. Early detection is critical for timely intervention, especially among high-risk clinical groups. Advances in wearable technology have recently enabled long-term, non-invasive cardiac monitoring through photoplethysmography (PPG), a light-based technique for tracking peripheral blood volume variations \cite{Backgroundpereira2020photoplethysmography}.
However, real-world PPG signals often suffer from motion artifacts, poor sensor contact, and physiological noise, making AF detection challenging \cite{motionaffectarunkumar2020heart, Backgroundpereira2020photoplethysmography}.

Existing AF detection approaches generally fall into two main categories. The first relies on handcrafted features derived from heart rate variability (HRV), such as the root mean square of successive differences (RMSSD) \cite{geurts2023heart, zhao2025validation}, the standard deviation of successive differences (SDSD)\cite{orini2023long}, and the percentage of NN intervals differing by more than 40 milliseconds (pNN40) \cite{eerikainen2017validating, zhao2025validation} . While physiologically interpretable, these features are highly sensitive to signal quality and often perform poorly in the presence of noise or irregular rhythms.
The second category involves deep learning methods that learn discriminative features directly from raw PPG signals \cite{bashar2019atrial, shen2019ambulatory, antiperovitch2024continuous}. Although such methods typically outperform traditional techniques in terms of accuracy, they suffer from several limitations: (1) they treat AF detection as a binary classification problem, thus failing to capture the full spectrum of clinically relevant arrhythmias; and (2) they rely exclusively on single-channel PPG data, which reduces robustness under motion-induced distortions \cite{shen2019ambulatory, antiperovitch2024continuous, shashikumar2017deep}. Additionally, (3) these models often discard low-quality signal segments, leading to inefficient utilization of continuous monitoring data \cite{bashar2019atrial, aliamiri2018deep, selder2020assessment, nguyen2022detecting}.

To address this, we propose a multimodal framework that combines PPG and ACC signals via a residual neural network augmented with temporal and channel attention mechanisms.
We evaluate it on a clinical-grade dataset of 1,000 hours of synchronized PPG, ECG, and ACC recordings from 49 elderly cardiac inpatients. To reflect real-world conditions, test samples are stratified by ACC-derived motion intensity without segment exclusion.
We compare RhythmiNet against a PPG-only deep learning model and a logistic regression baseline using HRV features, with performance evaluated across varying levels of motion intensity.

\section{METHOD}
\subsection{Dataset and Preprocessing}

\begin{figure*}[h]
\centering
\includegraphics[width=\textwidth]{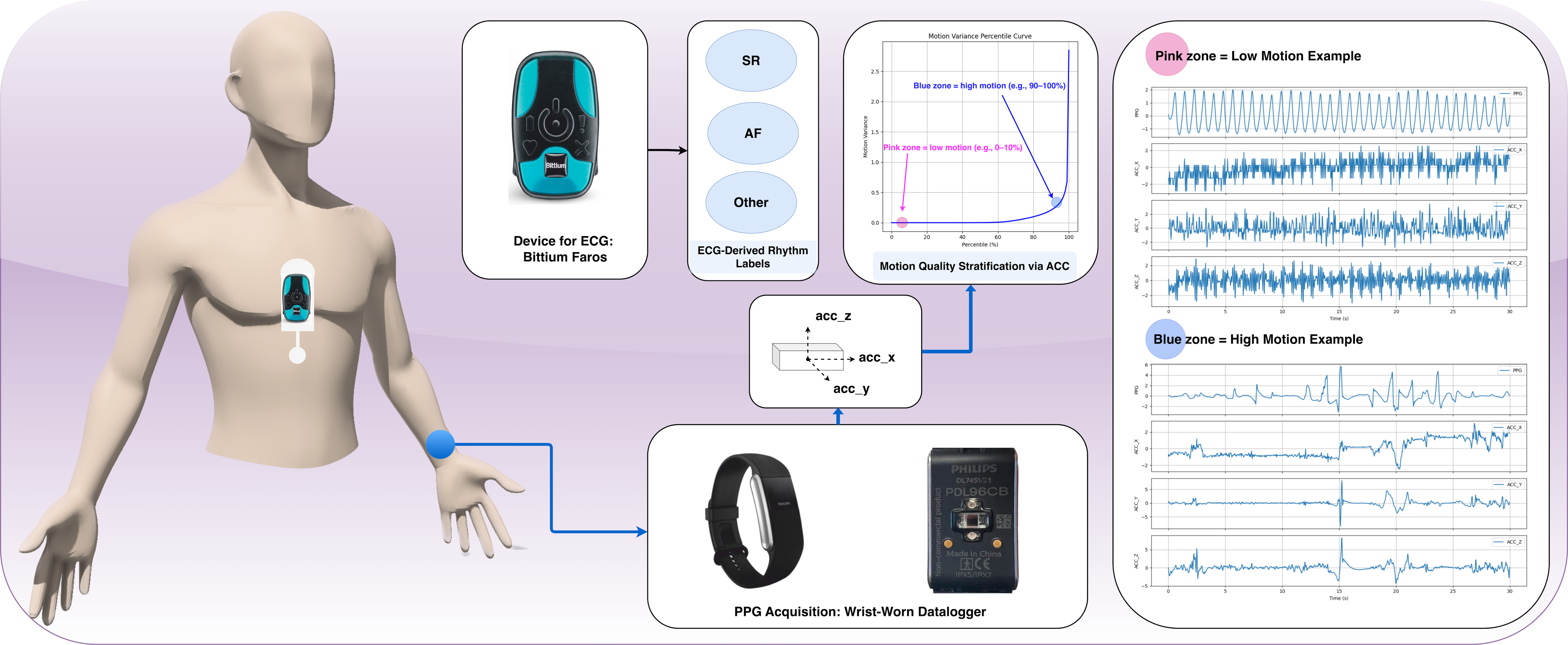}
\caption{
Overview of the data collection and motion stratification process. A wrist-worn Philips Datalogger recorded both PPG and tri-axial ACC signals, while a chest-worn Bittium Faros device recorded reference ECG. ECG provided rhythm labels (AF, SR, Other) used for training and testing. Segment-level motion scores were computed from the ACC magnitude and used to stratify test data into percentile-based motion levels. Example PPG and ACC traces are shown for two motion conditions: low-motion (top) and high-motion (bottom).}
\label{fig:flowchart}
\end{figure*}

\textbf{Data Collection:}  
This study uses a clinical-grade dataset collected between September 2022 and August 2023 from 49 elderly inpatients at the Heart Center of Turku University Hospital, Finland. Patients were monitored using two synchronized devices (Figure~\ref{fig:flowchart}): a wrist-worn Philips Datalogger (PPG and tri-axial ACC at 32 Hz) and a Bittium Faros\texttrademark~360 Holter (ECG at 125 Hz). Rhythm annotations (AF, SR, Other) were derived from the ECG signals. The "Other" category includes arrhythmias such as VT, IVR, arrest, COUP, GEM, and SALV, which do not fit into standard AF or SR classes. The dataset was split into patient-independent training (n = 25; 17 SR, 8 AF) and testing sets (n = 24; 18 SR, 6 AF), totaling 473.73 hours of PPG recordings for training and 461.73 hours for testing.

\textbf{Signal Preprocessing and Motion Stratification:}  
PPG signals were segmented into 30-second windows (960 samples at 32 Hz), followed by third-order Butterworth bandpass filtering (0.5–8 Hz) and normalization to zero mean and unit variance on each segment.
 ACC signals were segmented using the same temporal windows. Within each segment, raw tri-axial ACC data were bandpass-filtered (0.5–5.0 Hz, second-order Butterworth), and motion magnitude was computed as $\sqrt{acc_x(t)^2 + acc_y(t)^2 + acc_z(t)^2}$. Segment-level motion scores were defined as the variance of the magnitude signal \cite{al2023adaptive, aliamiri2018deep}. All test segments were ranked according to their motion scores and stratified into percentile bins (e.g., 0--10\%, 10--20\%, ..., 90--100\%) to evaluate model performance under varying levels of motion interference without excluding any data.

\subsection{Proposed Model Architecture}

\begin{figure*}[htbp]
    \centering
    \includegraphics[width=\textwidth]{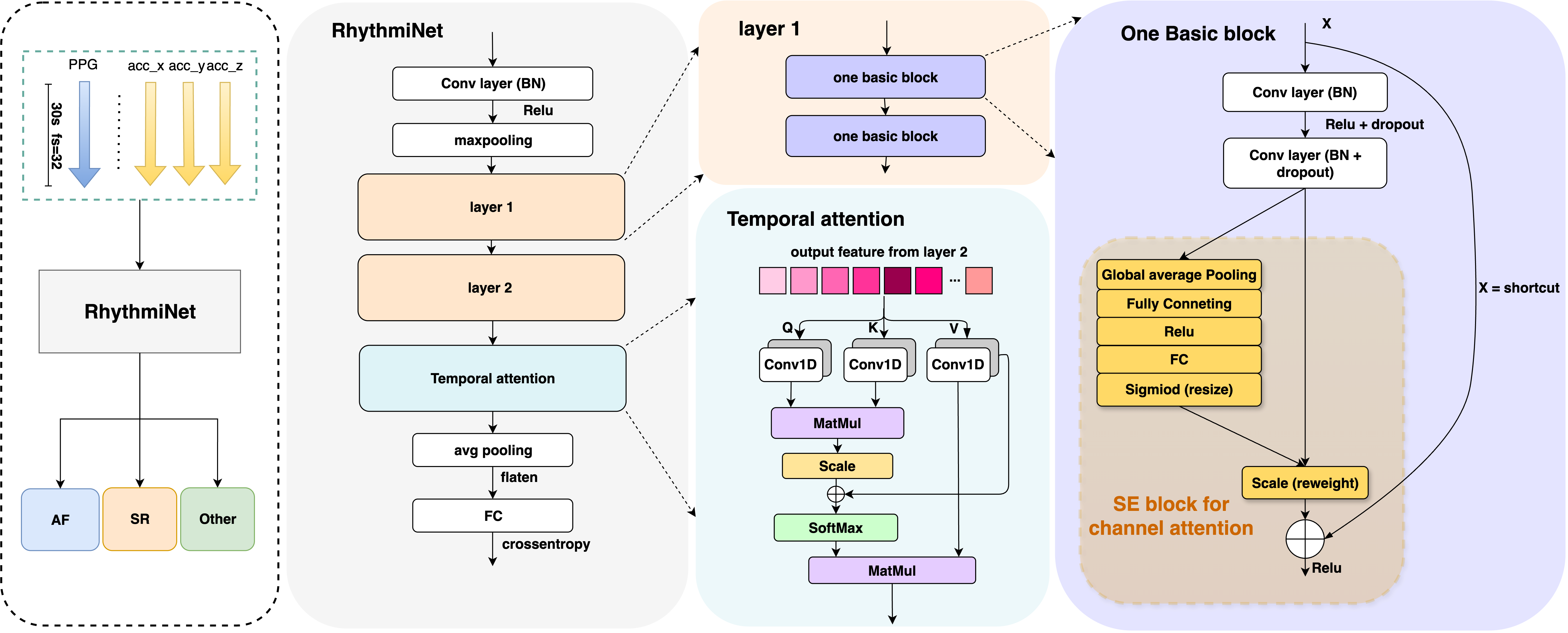}
    \caption{Overview of the RhythmiNet architecture for three-class heart rhythm classification (SR, AF, Other). The model processes 30-second segments of synchronized PPG and tri-axial ACC signals (960 samples at 32 Hz). Inputs are concatenated and passed through a convolutional stem, followed by two residual blocks enhanced with Squeeze-and-Excitation (SE) modules to emphasize informative channels. A temporal attention module captures long-range dependencies across time. Finally, global average pooling and a fully connected layer form the classification head. Dashed boxes illustrate the internal structure of selected modules (e.g., residual blocks and attention).
}
    \label{fig:network}
\end{figure*}

We introduce RhythmiNet, a lightweight residual neural network specifically designed to jointly process photoplethysmography (PPG) and tri-axial ACC signals for heart rhythm classification. As illustrated in Figure~\ref{fig:network}, the architecture is composed of four main components: (1) a convolutional stem, (2) a residual backbone \cite{he2016deep} enhanced with Squeeze-and-Excitation (SE) blocks \cite{hu2018squeeze}, (3) a temporal attention module \cite{vaswani2017attention}, and (4) a final classification head.
The model accepts four synchronized input channels—PPG and three-axis ACC—sampled over 30-second segments (960 samples at 32 Hz). The convolutional stem, equipped with stride-2 projection, reduces the temporal resolution and extracts low-level features. The residual backbone comprises two stages, each containing two SE-enhanced BasicBlocks. Each block consists of two 1D convolutional layers, batch normalization, dropout, and skip (residual) connections. The SE modules perform channel-wise reweighting by applying global average pooling followed by two fully connected layers with a reduction ratio of 2.
To model long-range temporal dependencies, a temporal attention module is appended after the second residual stage. This module implements scaled dot-product attention across the temporal dimension and integrates contextual information via residual connections.
Finally, the extracted features are aggregated using adaptive average pooling and passed through a fully connected layer. 

\subsection{Baseline Model}

We implemented a conventional HRV-based method as the baseline model. Specifically, three widely used time-domain features were extracted from inter-beat intervals (IBIs): the RMSSD \cite{geurts2023heart, zhao2025validation}, pNN40 \cite{orini2023long, zhao2025validation}, and the SDSD \cite{eerikainen2017validating}.
Pulse peaks were then detected using the Automatic Multiscale Peak Detection (AMPD) algorithm \cite{scholkmann2012efficient}. IBIs were computed as the time differences between adjacent peaks, and the HRV features were subsequently derived for each segment using standard analytical formulas.
Each of the three features was used independently to train a logistic regression classifier. This classical approach serves as a practical baseline.

\subsection{Experimental Setup}

The model was trained on the full train dataset using three random seeds for robustness. Each run used a batch size of 64 for 60 epochs. To reduce temporal overfitting, signal segments were shuffled at each epoch. Optimization used Adam (learning rate \(1 \times 10^{-4}\), weight decay \(1 \times 10^{-5}\)) with a StepLR scheduler (factor 0.1 every 20 epochs). Cross-entropy loss was used for training.
Experiments were conducted on a Linux workstation with Python 3.11.9, PyTorch 2.3.0, and an NVIDIA TITAN RTX GPU (24 GB, CUDA 12.1). 
We report macro-AUC, micro-AUC, and accuracy. Macro-AUC treats all classes equally, while micro-AUC reflects true label distribution. Accuracy denotes the proportion of correct predictions.

\section{EXPERIMENT RESULTS}

\begin{figure*}[htbp]
    \centering
    \includegraphics[width=\textwidth]{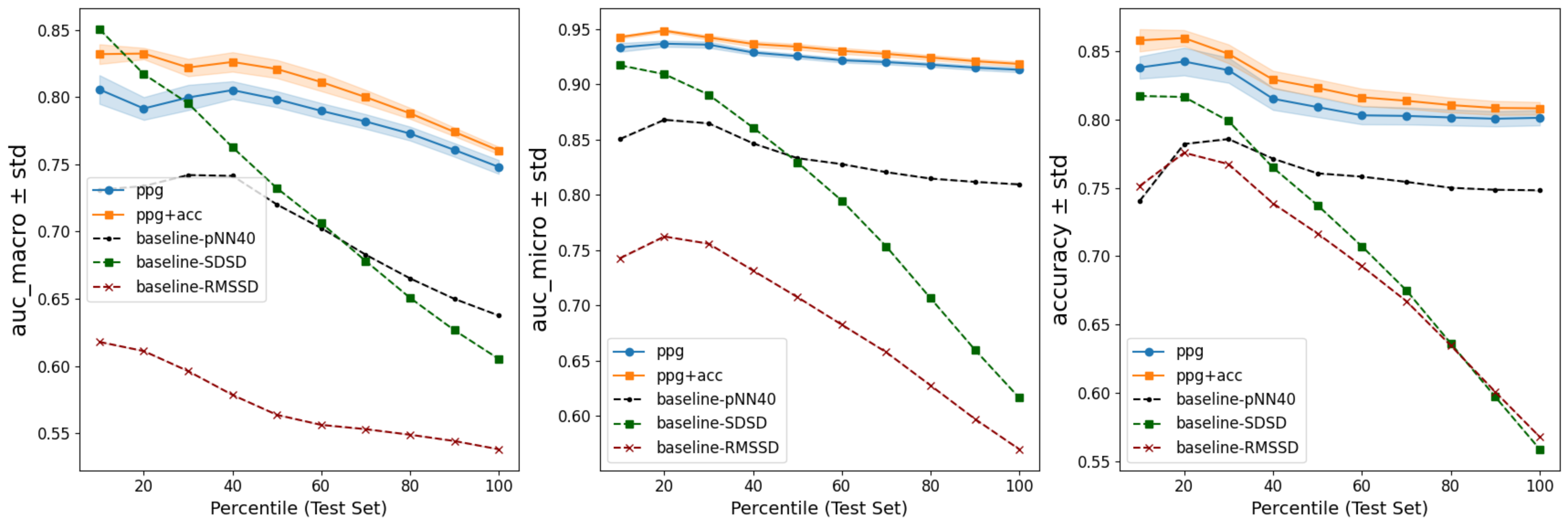}
    \caption{Performance across motion intensity percentiles: (Left) Macro-AUC, (Middle) Micro-AUC, and (Right) Accuracy.}
    \label{fig:performance_percentile}
\end{figure*}

Figure~\ref{fig:performance_percentile} presents a comparison of model performance across motion percentiles using three evaluation metrics: macro-AUC, micro-AUC, and accuracy. The proposed model using combined PPG and ACC input (orange) consistently outperforms all other methods across all metrics and motion levels. Specifically, it achieves a macro-AUC exceeding 0.83 in low-motion conditions (0–10\%) and maintains robust performance (around 0.76) even in high-motion segments (90–100\%). 

In terms of micro-AUC, the PPG+ACC model remains consistently above 0.91, highlighting its strong generalization ability despite motion interference. The PPG-only variant (blue) also shows competitive performance, though with a more noticeable decline under increased motion. Accuracy follows a similar trend: the PPG+ACC model maintains above 0.81 across all percentiles, while the PPG-only model dips slightly but stays above 0.80.

In contrast, all HRV-based baselines, including pNN40, SDSD, and RMSSD, show steep performance degradation as motion intensity increases. Both SDSD and RMSSD fall below 0.60 in macro-AUC and accuracy in the top motion percentile, indicating their vulnerability to motion artifacts and limited robustness in realistic wearable settings.

\section{DISCUSSION AND FUTURE WORK}

RhythmiNet improves AF detection by leveraging multimodal inputs—PPG and ACC—to enhance robustness in motion-intensive and noise-prone conditions. Instead of relying on complex fusion strategies, the model simply concatenates the input channels and processes them through a compact residual backbone. Temporal attention and SE modules assist in extracting relevant features without adding complexity. We hypothesize that the model’s robustness to motion interference arises from three factors: (1) the inclusion of ACC signals to provide motion context, (2) attention-based reweighting of noisy temporal segments, and (3) inclusive training using motion-contaminated data without exclusion.
This effective design outperforms traditional PPG-only methods and demonstrates strong potential for deployment in wearable cardiac monitoring.

To further enhance the clinical applicability of RhythmiNet, future work will focus on three main directions. First, we aim to improve model interpretability by developing intuitive visualization tools that highlight which input segments most influence decision-making. Second, we will investigate model compression and acceleration techniques, such as pruning and quantization, to support real-time inference on low-power, resource-constrained wearable devices. Finally, we intend to benchmark RhythmiNet against a wider range of state-of-the-art models across diverse, publicly available datasets to comprehensively evaluate its generalizability and clinical utility.

\section{ACKNOWLEDGMENTS}
This study was part of the clinical trial CARE-DETECT (ClinicalTrials.gov ID: NCT05351775). It was approved by the Ethics Committee of the Hospital District of Southwest Finland and conducted in accordance with the Declaration of Helsinki. Written informed consent was obtained from all participants. 

This study was funded by the Moore4Medical project, supported by the ECSEL JU and Business Finland (Grant Agreements H2020-ECSEL-2019-IA-876190 and 7215/31/2019), and by the ITEA project RM4HEALTH, supported by Business Finland (Grant 8139/31/2022).

\bibliographystyle{ACM-Reference-Format}
\bibliography{sample-base}

\end{document}